# Hierarchical energy signatures using machine learning for operational visibility and diagnostics in automotive manufacturing

Ankur Verma [a], Seog-Chan Oh [b], Jorge Arinez [b], Soundar Kumara [a]

[a] Harold and Inge Marcus Department of Industrial and Manufacturing Engineering, The Pennsylvania State University, University Park, PA, USA
[b] Manufacturing Systems Research Lab., General Motors R&D, Warren, MI, USA

Abstract

Manufacturing energy consumption data contains important process signatures required for operational visibility and diagnostics. These signatures may be of different temporal scales, ranging from monthly to sub-second resolutions. We introduce a hierarchical machine learning approach to identify automotive process signatures from paint shop electricity consumption data at varying temporal scales (weekly and daily). A Multi-Layer Perceptron (MLP), a Convolutional Neural Network (CNN), and Principal Component Analysis (PCA) combined with Logistic Regression (LR) are used for the analysis. We validate the utility of the developed algorithms with subject matter experts for (i) better operational visibility, and (ii) identifying energy saving opportunities.

## 1. Introduction

The automotive industry is moving towards sustainable practices and reducing energy usage, for which better operational visibility is the first step [1, 2, 3]. Globally, manufacturing is the largest sector responsible for greenhouse gas emissions accounting for a mammoth 31% of the 51 billion tons of emissions [4]. General Motors has an annual global energy consumption of 9000 Gigawatt hours (GWh), roughly equivalent to a mid-sized city's annual energy consumption (1 GW power consumption) [5]. With low-cost sensors, energy meters, and Internet of Things (IoT) capabilities, there is an increased adoption of measurement technologies oriented towards an energy-conscious future.

Machine learning and deep learning techniques are increasingly being investigated in addition to process-based techniques due to (i) availability of large amounts of sensor data, (ii) self-improvement capabilities of neural networks, (iii) and the ability to re-use models trained at one temporal scale or manufacturing location to another (transfer learning). Previously, researchers have investigated supervised and unsupervised machine learning (ML) techniques for studying the energy profiles of specific machines and manufacturing processes [6, 7], and improving energy efficiency in industry [8-15]. Energy consumption data can also be used to infer manufacturing process signatures [16-20]. Process signatures in this context refer to the energy consumption patterns of operations like running a motor, pump, ovens, etc. These patterns can then be used to identify operational states like normal, abnormal, and maintenance.

However, in many industrial environments, electric lines are not mapped to specific processes, making a process-electricity map a pre-requisite for any further investigation. Furthermore, the collected energy data may have periodicity at multiple-temporal scales like weekly or monthly recurrence, and identifying a specific event may require parsing through years of time series data. Moreover, each facility may have a different set of machines and operating conditions, requiring approaches like transfer learning to deploy and scale machine learning models.

In this study, we investigate the energy signatures of the Pre-Treatment and Electro-Dip (PTED) section of the paint shop, which is one of the more energy-intensive processes in automotive manufacturing. Our key contributions are: (i) a process-electricity map for inferring process energy consumption from 20-30 electric lines typical in a manufacturing unit, where each process comprises of several machines, (ii) enhanced operational visibility of the PTED process at different temporal scales like yearly, weekly, monthly, and daily, and (iii) transfer learnable labeling and diagnostics using different machine learning techniques.

We first develop a transferable process-electricity map for the PTED process in the paint shop of automotive manufacturing for mapping electric lines to specific process equipment, shown in Fig. 1. We use three techniques in this work: a Multi-Layer Perceptron (MLP) and a Convolutional Neural Network (CNN) for supervised learning and Principal Component Analysis + Logistic Regression (PCA + LR) for unsupervised learning. For supervised learning, we use a two-stage hierarchical network structure to narrow down the temporal search space. Each stage of the network deals with temporal data at different scales: weekly and daily. We generate labeled data for different operational conditions as defined by subject matter experts. Depending on the temporal scales, they are as follows: (i) for weekly analysis: normal, higher, anomaly, and start issues, and (ii) for daily analysis: production, maintenance and startup, shutdown, maintenance, and anomaly. The difference between start issues and anomalies is that start issues happen while ramping up of the systems whereas anomalies occur when the system is already operational and fluctuates from the normal operating load.

The two-stage approach will help process engineers narrow down the search space and focus on weeks showing interesting 'electricity signatures'. Though the energy consumption patterns are substantiated using production numbers, we are only using energy data in this manuscript due to the proprietary nature of production data.

## 2. Methodology

This section details the methodology used in this work to (i) collect and pre-process the data, (ii) perform supervised learning using an MLP and a CNN, and (iii) perform unsupervised learning using a combination of PCA+LR [21, 22].

### 2.1 Data collection and pre-processing

The PTED section in the paint shop involves several processes like immersion, spray, rinsing, and ovens, each of which require different electromechanical components. To gain the process insight from electricity data, the first requirement is to have a process-electricity mapping. However, the electric wires are not set up and distributed according to the process to which they supply, which makes this mapping a pre-requisite engineering task. Fig.1. shows a part of the process-electricity map prepared with the help of domain experts to understand the processes (immersion, pump, ash house, etc.) corresponding to the specific electricity lines (LINE-1, LINE-2, etc.).

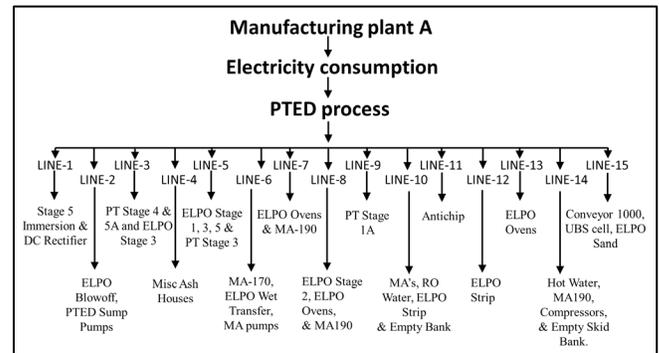

Fig.1. PTED process-electricity map for the studied factory

### 2.2 Hierarchical decomposition for weekly and daily analysis

The two-stage hierarchical decomposition approach for weekly and daily analysis is shown in Fig.2. We start with long term (3 years) data and then extract a 3-month data chunk from the data lake. This is because it is easier for subject matter experts to deal with a specific problematic chunk of data rather than all the raw data. The class labels generated with the help of domain experts and hierarchical breakdown from 3 year – 3 month – 1 week of data are shown in Fig.2. below.

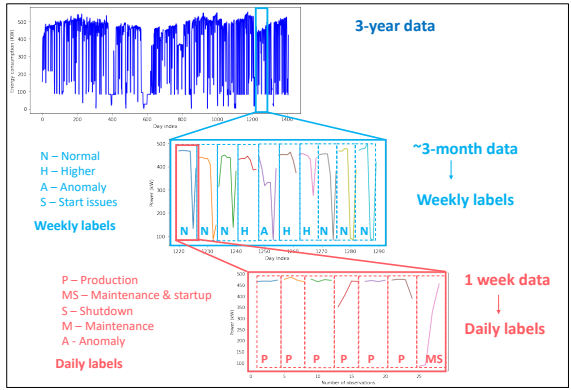

Fig.2. Extracting and labelling monthly and weekly data from a 3-year data chunk

## 2.3 Neural network modeling

We use a two-stage hierarchical approach for both: an MLP and a CNN for performing weekly and daily analysis on the PTED electricity data. Fig.3. shows the two-step hierarchical network for the MLP implementation, along with the input and output data types and classes labels.

### 2.3.1 Stage 1: Weekly analysis and labelling

The stage 1 network is a (i) 3-layer deep MLP, or a (ii) CNN with 2 convolutional layers followed by maxpooling, dropout, flatten, and dense layers, which take 7 data points per week as input and classify the weekly production into one of the following four classes: (i) normal, (ii) higher, (iii) anomaly, or (iv) start issues.

### 2.3.2 Stage 2: Daily analysis and labelling

The stage 2 network is a (i) 3-layer deep MLP, or a (ii) CNN with 2 convolutional layers followed by maxpooling, dropout, flatten, and dense layers, which take 4 data points per day as input and classify the daily production into one of the following five classes: (i) production, (ii) maintenance & startup, (iii) shutdown, (iv) maintenance, and (v) anomaly.

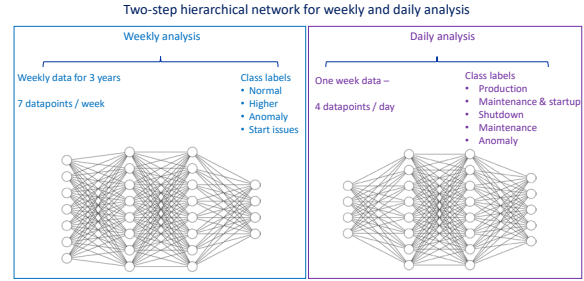

Fig.3. Two-step hierarchical network for weekly and daily analysis

## 2.4 Principal component analysis (PCA) + Logistic Regression (LR) modeling

We use a pipeline of PCA and LR to obtain energy signatures in an unsupervised way. The PCA provides an unsupervised dimensionality reduction to mitigate the issue of multicollinearity (high dependence) among the explanatory variables, while the logistic regression does the prediction based on the principal components represented by Eigenvectors found in the PCA. 5 Eigenvectors represent the weekly analysis data in this work. This is implemented in python using the following classes from Scikit-learn, a popular machine learning library:
(i) sklearn.decomposition.PCA
(ii) sklearn.linear_model.LogisticRegression
(iii) sklearn.pipeline.Pipeline
(iv) sklearn.grid search.GridSearchCV

## 3. Results

We provide the results of the three tested models below. The test accuracies of weekly and daily analyses on a particular electric line data (LINE-2 in Fig. 2) are shown in Fig.4.

### 3.1 Model validation

This section evaluates the three models developed in this study by using the cross-validation (CV) technique, as follows:
1. Divide the dataset into two parts: 75% data for training, 25% for testing, where a random seed number is specified to ensure repeatability.
2. Train the model on the training set.
3. Validate the model on the test set.

4. Repeat 1-3 steps 100 times as the seed number changes.

Fig.4(a, b). show the cross-validation results on 100 trials of weekly and daily level test data sets, respectively. In weekly level data set validation, the average accuracy of MLP is 92%, CNN is 95%, and PCA+LR is 94%. Meanwhile, in daily level data set validation, average accuracy of MLP is 92%, CNN is 97%, and PCA+LR is 95%. The authors believe that the reason CNNs outperform other models is due to their (i) local receptive fields, (ii) shared weights, and (iii) spatial or temporal subsampling [23, 24].

### 3.2 Weekly and daily analysis results

Fig.4(a, b). shows the test accuracies of weekly and daily analysis using MLP, CNN, and PCA+LR models.

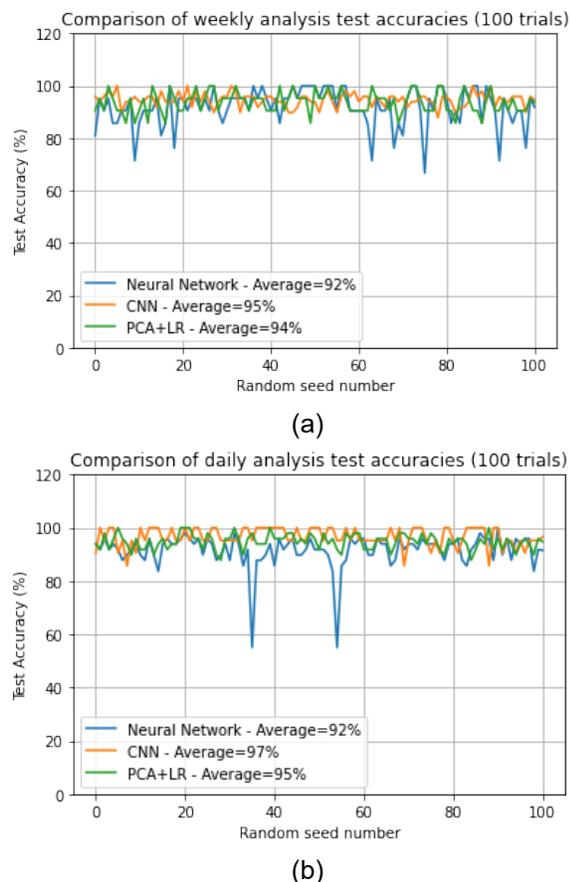

(a)

(b)

Fig.4. Test accuracies using MLP, CNN, and PCA+LR for (a) weekly and (b) daily analysis.

### 4. Conclusions

We identify automotive manufacturing process signatures from electricity data using three machine learning techniques: MLP, CNN, and PCA+LR. Automated generation of weekly and daily labels from electricity data about the PTED process helps reduce the search space to investigate specific events of interest. The results show that the CNN performs the best, followed by PCA+LR, followed by the MLP, which is 2-3% less accurate than the PCA+LR model. The models developed in this study only focused on electricity data from the PTED shop. The energy source (natural gas for heating, etc.) and the type of production will determine the applicability of specific machine learning models. There are three interesting directions for future work: (i) using alternative architectures like the transformer neural network (ii) using unsupervised learning approaches specifically for clustering as a steppingstone to supervised learning, and (iii) using cumulative power or energy for labeling and diagnostics.

### Declaration of Competing Interest

The authors declare that they have no known competing financial interests or personal relationships that could have appeared to influence the work reported in this paper. SK acknowledges Allen E. Pearce and Allen M. Pearce Professorship for their generous support for this work.

### References

[1] Verma A, Goyal A, Kumara S, Kurfess T, Edge-cloud computing performance benchmarking for IoT based machinery vibration monitoring, Manuf Lett, 2021; 27:39-41.
[2] Zamorano J, Alfaro M, Martins de Oliveira V, Fuertes G, Durán C, Ternero R, Sabattin J, Vargas M, New manufacturing challenges facing sustainability, Manuf Lett, 2021; 30:19–22.
[3] Verma A, Sharma M, Developing An understanding of industry 4.0 using automotive manufacturing as a reference, FISITA World Automotive Congress 2018, Chennai, India.
[4] https://breakthroughenergy.org/our-approach/the-data/#circles, Accessed January 19th; 2023.
[5] https://www.edisonenergy.com/case-study/general-motors-and-the-road-to-100-renewable-

energy/#:~:text=With%20an%20annual%20global%20energy,GM's%20energy%20portfolio%20is%20immense, Accessed January 19th; 2023.

[6] Tan D, Suvarna, M, Shee Tan Y, Li J, Wang X. A three-step machine learning framework for energy profiling, activity state prediction and production estimation in Smart Process Manufacturing. Applied Energy. 2021;291:116808.

[7] Thiede S, Turetskyy A, Loellhoeffel T, Kwade A, Kara S, Herrmann, C. Machine learning approach for systematic analysis of energy efficiency potentials in manufacturing processes: A case of battery production. CIRP Annals 2020;69(1):21–24.

[8] Narciso DAC, Martins FG. Application of machine learning tools for energy efficiency in industry: A Review. Energy Reports 2020;6:1181–1199.

[9] Chicco G. Overview and performance assessment of the clustering methods for electrical load pattern grouping. Energy 2012;42(1):68–80.

[10] Zhou K, Yang S, Shen C. A review of electric load classification in Smart Grid Environment. Renewable and Sustainable Energy Reviews 2013; 24:103–110.

[11] Miller C, Nagy Z, Schlueter A. A review of unsupervised statistical learning and visual analytics techniques applied to performance analysis of non-residential buildings. Renewable and Sustainable Energy Reviews. 2018;81:1365–1377.

[12] Oh S-C, Hildreth AJ. Decisions on energy demand response option contracts in smart grids based on activity-based costing and stochastic programming. Energies 2013;6(1):425-443.

[13] Oh S-C, Hildreth AJ. Estimating the technical improvement of energy efficiency in the automotive industry – stochastic and deterministic frontier benchmarking approaches. Energies. 2014;7(9):6198-6222.

[14] Huang A, Badurdeen F. Metrics-based approach to evaluate sustainable manufacturing performance at the production line and Plant Levels. Journal of Cleaner Production 2018, 192, 462–476.

[15] Zhang H, Haapala K. R. Integrating Sustainable Manufacturing Assessment into decision making for a production work cell. Journal of Cleaner Production. 2015, 105, 52–63.

[16] Yang H, Kumara S, Bukkapatnam STS, Tsung F. The internet of things for smart manufacturing: a review. IISE Trans 2019;51(11):1190–216.

[17] Mohanty AR. Machinery Condition Monitoring: Principles and Practices. CRC Press, Taylor and Francis Group; 2014.

[18] Oh S-C, Hildreth AJ. Analytics for smart energy management. Springer; 2016.

[19] https://www.energy.gov/sites/default/files/2022-06/2018_mecs_all_manufacturing_sankey.pdf, Accessed January 19th; 2023.

[20] Quadrennial Technology Review, 2015, https://www.energy.gov/sites/prod/files/2017/03/f34/quadrennial-technology-review-2015_1.pdf, Accessed January 19th; 2023.

[21] Rumelhart DE, Hinton GE, Williams RJ. Learning representations by backpropagating errors. Nature. 1986;323(6088):533–536.

[22] Goodfellow I, Bengio Y, Courville A. Deep learning. MIT press; 2016.

[23] Lecun Y, Bottou L, Bengio Y, Haffner P, Gradient-based learning applied to document recognition. Proceedings of the IEEE. vol. 86, no. 11, pp. 2278-2324, Nov. 1998, doi: 10.1109/5.726791.

[24] Simonyan K, Zisserman A, Very Deep Convolutional Networks for Large-Scale Image Recognition, arXiv e-prints, 2014. doi:10.48550/arXiv.1409.1556.